\documentclass[10pt,twocolumn,letterpaper]{article}

\usepackage{cvpr}
\usepackage{times}
\usepackage{epsfig}
\usepackage{graphicx}
\usepackage{amsmath}
\usepackage{amssymb}
\usepackage{float}
\usepackage{caption}
\usepackage{subcaption}
\usepackage{authblk}

\usepackage[pagebackref=true,breaklinks=true,letterpaper=true,colorlinks,bookmarks=false]{hyperref}

\usepackage{pifont}
\cvprfinalcopy 

\ifcvprfinal\fi
\begin{document}

\title{Pyramid Point: \\A Multi-Level Focusing Network for Revisiting Feature Layers}

\author[]{Nina Varney}
\author[]{Vijayan K. Asari}
\author[]{Quinn Graehling}
\affil[]{Department of Electrical Engineering, University of Dayton\\
{\tt\small \{varneyn1, vasari1, graehlingq1\}@udayton.edu}}
\renewcommand\Authands{ and }


\maketitle

\begin{abstract}
We present a method to learn a diverse group of object categories from an unordered point set. We propose our Pyramid Point network, which uses a dense pyramid structure instead of the traditional 'U' shape, typically seen in semantic segmentation networks. This pyramid structure gives a second look, allowing the network to revisit different layers simultaneously, increasing the contextual information by creating additional layers with less noise. We introduce a Focused Kernel Point convolution (FKP Conv), which expands on the traditional point convolutions by adding an attention mechanism to the kernel outputs. This FKP Conv increases our feature quality and allows us to weigh the kernel outputs dynamically. These FKP Convs are the central part of our Recurrent FKP Bottleneck block, which makes up the backbone of our encoder. With this distinct network, we demonstrate competitive performance on three benchmark data sets. We also perform an ablation study to show the positive effects of each element in our FKP Conv. 
\end{abstract}

\begin{figure}[t]
\begin{center}
\includegraphics[width=0.99\linewidth]{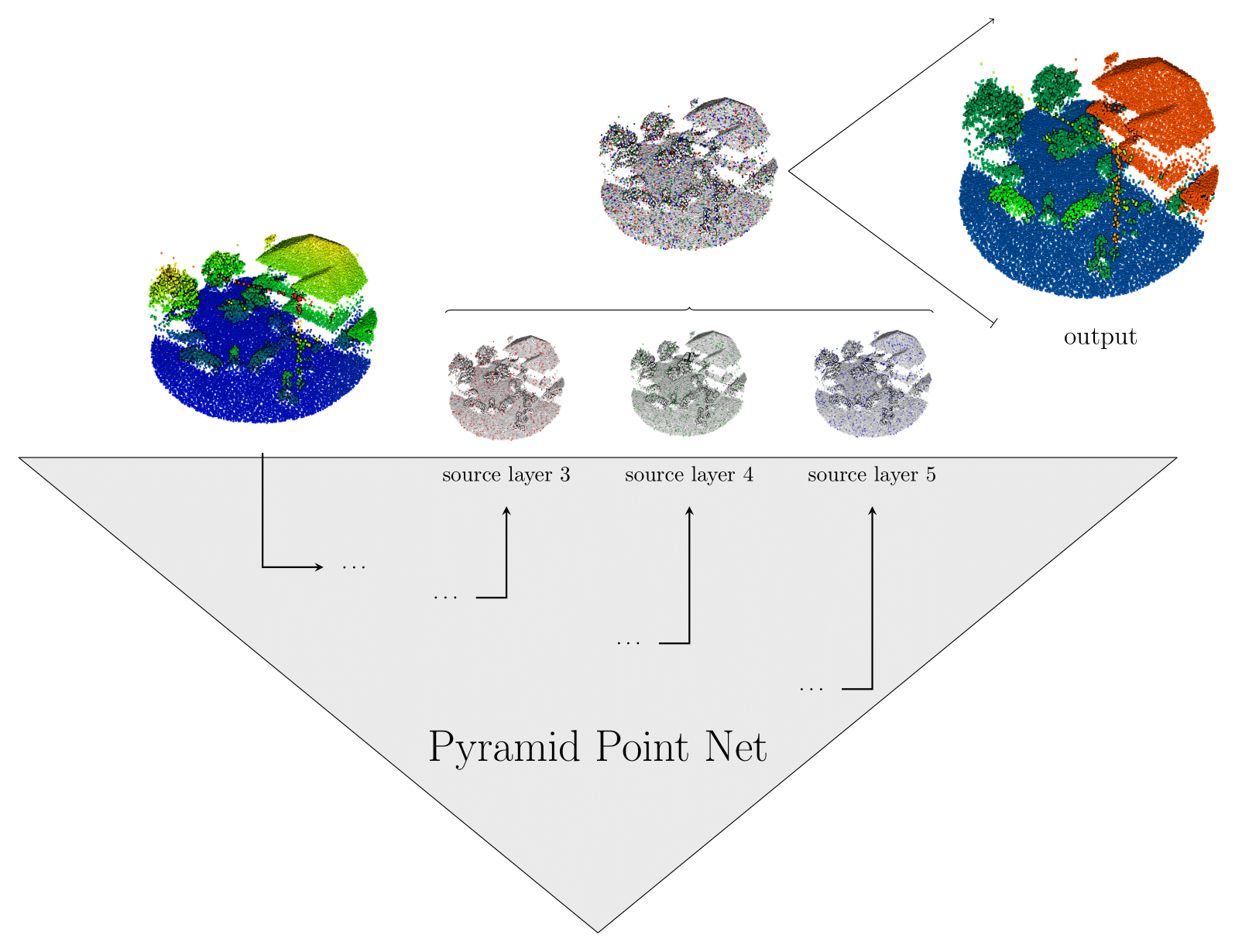}
\end{center}
   \caption{Example of the fundamental concept of our Pyramid Point network. The final output is based not only on an upsampling from the previous layers; but also on features from all network layers. By upsampling from each layer and then concatenating the features, we can get a richer combination with less noise.}
\label{fig:onecol}
\end{figure}

\section{Introduction}

Recent breakthroughs in automatic driving and the decrease in LiDAR sensor cost have increased research into 3D object detection. One of the most popular tasks is the semantic segmentation of unorganized 3D point clouds. 3D point clouds present a distinct type of challenge because of their sparse and unordered nature. These point clouds exist as lists of points containing X, Y, and Z information with other possible features like reflectively or RGB data. The resolution and occlusions in a point cloud vary greatly, not only from sensor-to-sensor, but also within each scene. The points are spatially related, however the relationships are not uniform like we would find in 2D imagery.    This property makes operations like convolution difficult.

There are two main approaches to handling convolutions with unordered point sets. The first is to transform the point clouds into ordered sets, either through voxels or transformations, into a higher dimension. The second approach is to redefine the convolution operation to perform on unordered point sets.  Once a convolution operation is defined, we can begin to apply similar methods, like those used in 2D scene segmentation. Most 3D scene segmentation methods use the traditional U-Net type structure with encoder layers, followed by decoder layers, usually with connections between similar sized encoder and decoder layers. 
However, this U structure has some shortcomings, particularly rough results that struggle to segment fine details. This failure is due to the max-pooling and subsampling of the feature layers, which results in a reduced feature map resolution. This extreme reduction in the receptive field at each encoder layer makes it very difficult for this network structure to segment small objects with a high degree of accuracy. 

We propose our Pyramid Point network, which uses feature fusion to combat reduced feature map resolution issues. Pyramid Point passes features in a dense pyramid structure. This structure results in feature fusion, from the encoding layers to the corresponding decoding layers, between encoder and decoder units of different layers and between the decoder layers themselves. By allowing the features to traverse between different units and layers, not just restricting them to the defined U path, we will enable the network to gain different receptive field views. The network output layer is in Figure \ref{fig:onecol}. This structure also provides the advantage of having several "shallow" feature layers, which have not gone through the entire U structure. Because the decoding unit adds noise, features that have gone through one or two decoder units, instead of the traditional four, are less susceptible to segmentation errors, which is especially crucial for small objects. 

In addition to the proposed pyramid structure, we also introduce the Focused Kernel Point Convolution (FKP Conv), which adds an attention element to the kernel outputs of a traditional kernel point convolution. This FKP Conv is the critical element in our Recurrent FKP Bottleneck block. 

In our evaluation section, we show the success of our semantic segmentation network. We assess our performance on three different networks, DALES \cite{dales}, Paris-Lille 3D \cite{npm3d}, and Semantic 3D \cite{semantic3d}. We choose these three LiDAR data sets because of their distinctly different signatures, aerial, mobile, and terrestrial type LiDAR, respectively. This evaluation strategy allows us to assess our network's overall performance and robustness to different object categories, resolutions, and occlusions. 
We show that our network outperforms other networks on the DALES and Paris-Lille 3D data sets while providing competitive performance on the Semantic 3D data set and lessening the variation in results across object categories. Finally, we perform an ablation study to assess the FKP Conv elements and their effect on our overall performance. Specifically, we explore the impact of the number of hidden layers in the recurrent FKP Conv and the implications of different pooling types. 

Our contributions are the following:
\begin{itemize}
\itemsep-0.2cm
\item We propose our Pyramid Point network, a multi-level focusing network which allows for dense feature fusion across all layers of the network
\item We introduce a Focused Kernel Point Convolution (FKP Conv) which adds spatial and channel attention elements to the traditional kernel point convolutions
\item We demonstrate the effectiveness of this network on three  benchmark data sets, representing three distinct LiDAR types
\end{itemize}

\section{Related Works}

\subsection{Semantic Segmentation of Point Clouds}
Since AlexNet's \cite{alexnet} debut in 2012, there has been an overwhelming amount of research in developing deep convolutional networks for 2D imagery. As this field continues to experience massive success, some focus has shifted into applying deep learning to other types of data, specifically 3D imagery. Developing deep learning methods for 3D imagery becomes increasingly difficult when the data is unstructured, as is the case in mesh and point cloud data. 

In point cloud data, many original methods focused on transforming the point cloud into a structured data set using voxels, such as VoxNet \cite{voxnet}, VoxelNet \cite{voxelnet}, and Point-Voxel CNN \cite{pointvoxelcnn}. In terms of transformation, voxels do the best job of maintaining fidelity to the point cloud's original shape. However, voxels present an issue because they require that the point cloud be sampled at a set size typically chosen by the user.

Recently, newer methods have concentrated on performing operations directly on the points themselves. PointNet \cite{pointnet} was the first architecture to use point sets as inputs, and the authors developed a method to extract global features from these point sets. PointNet++ \cite{pointnet++} expanded on this by adding a local hierarchy. In addition to the local hierarchy, PointNet++ also introduced the idea of multi-scale grouping and multi-resolution grouping to deal with the problems of different scales in varying resolutions within a single scene. PointNet++ popularized the idea of performing the operations in a non-Euclidean metric space, such as using geodesic distances, to capture the underlying points instead of the traditional metric space. Architectures such as VoteNet \cite{votenet}, ShellNet \cite{shellnet}, PointCNN \cite{pointcnn}, and  SplatNet \cite{splatnet} expand on using 3D points as direct inputs. 

Pointwise multi-layer perceptron networks, like PointNet, were prevalent, but several networks have recently defined specific point convolutions performed directly on the inputs. PointwiseCNN \cite{pointwise}, SpiderCNN \cite{spidercnn}, flex convolution \cite{flexconvolution}, and PCNN \cite{pcnn} all focus on defining these point convolution operations.  The most notable point convolution operation is the Kernel Point Convolution (KP Conv) \cite{kpconv}, which uses a linear correlation to connect the kernel points to their closest input points. The authors also introduce the idea of a deformable convolution, where kernel points can adapt to local geometry. 

\begin{figure*}[htbp!]
\begin{center}
\includegraphics[width=0.99\linewidth]{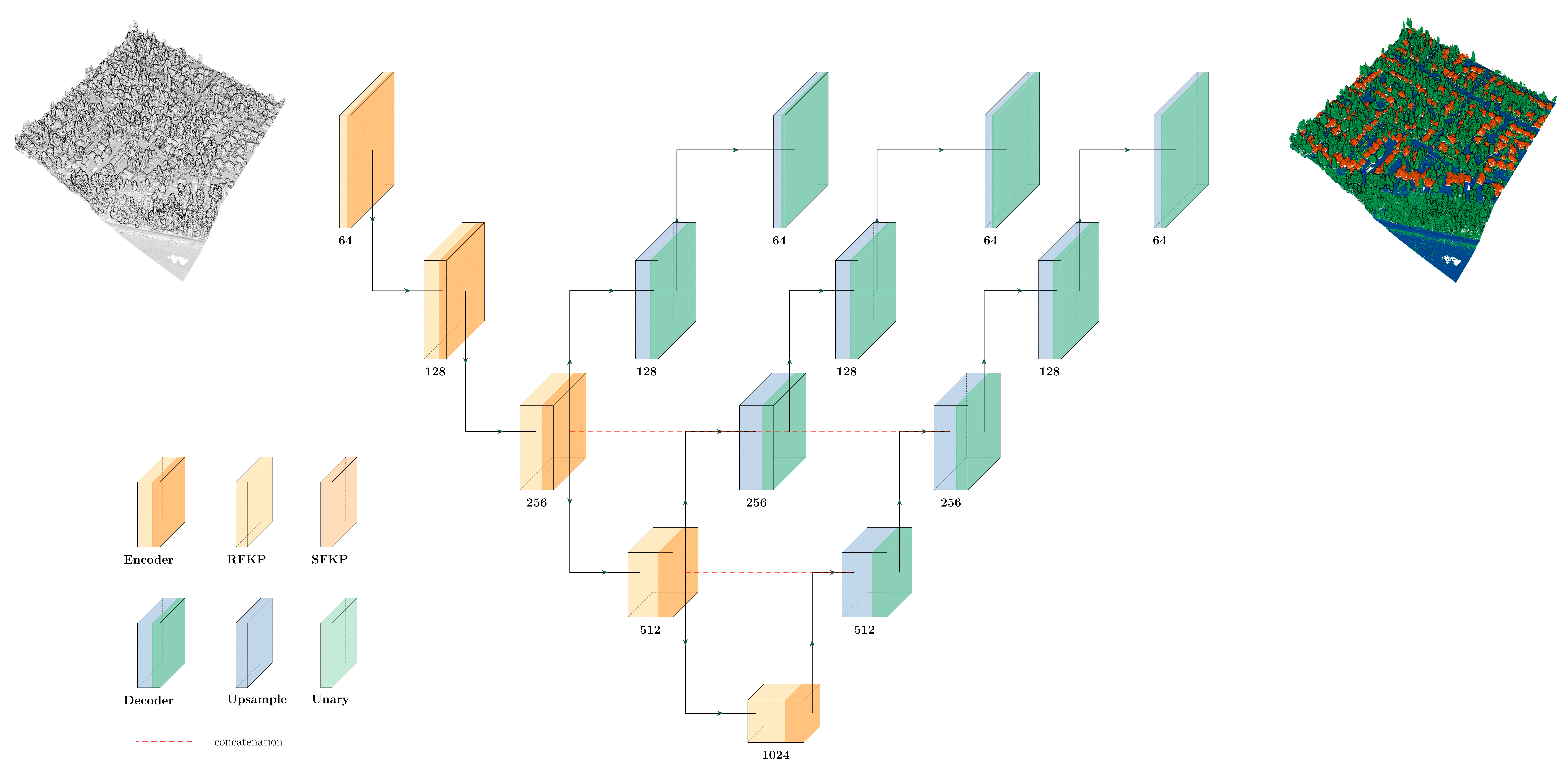}
\end{center}
   \caption{Example of the architecture of our Pyramid Point network. We construct the architecture in an inverse dense pyramid shape instead of the typical U shape. At the third layer, features are both downsampled and upsampled, and similarly, dimensioned layers are concatenated together in the final stages. }
\label{fig:arch}
\end{figure*}

\subsection{Pyramid Networks}
Features used for computer vision application have long relied on scaling to provide rich and robust features. Classical computer vision methods such as Scale Invariant Feature Transforms (SIFT) \cite{lowe2004sift} and Speeded Up Robust Features (SURF) \cite{bay2006surf} rely heavily on these methods. This approach has also shifted into deep learning, where architectures rely on a U-type structure to perform image segmentation and scene understanding. 

Medical image segmentation tasks have a rich history of hierarchical methods. U-Net \cite{unet} is the prominent architecture, suggesting a hierarchical approach to image segmentation with cross-architecture connections. From there, many iterations such as Residual U-Net, R2 U-Net \cite{r2unet}, and SegNet \cite{segnet} made incremental improvements. Finally, NABLA-N ($\nabla^{N}$) Net \cite{nablan} proposed a pyramid structure of hierarchical layers. While many improvements to the typical encoder-decoder structure have been made in the 2D image space, the 3D image space has mostly the same architectures, focusing on improvements to the convolutional units themselves rather than the architecture. 

We propose a 3D semantic segmentation network that moves away from the U structure in favor of a pyramid network. Each layer is encoded and decoded at multiple points within the network. This approach gives us multiple viewpoints at different hierarchical levels and allows us to access different scales while avoiding the noise associated with traversing through the entire architecture. 

\subsection{Attention Modules}
Attention modules have significantly impacted deep learning with their ability to efficiently model dependencies \cite{attentivesurvey, attnsurvey}. Originally developed to combat long-range dependence in encoder-decoder-based natural language processing methods, attention has expanded into many deep learning areas, including computer vision. Recent proposals to the self-attention mechanism have expanded the scope to include scene segmentation tasks \cite{interlaced}. Modern computer optics allow for image captures to be represented by complex and densely packed features. The pixels representing a specific label's features are often susceptible to changes in various fields, such as scaling and rotation, leading to a lack of consistency within a semantic class. The attention module can capture rich contextual relationships for better feature representation, which helps alleviate intra-class conflicts and improves overall segmentation accuracy. 

Attention mechanisms have many versions, but the most popular lately are varying channel and spatial attention combinations. RCAN \cite{rcan}, CBAM \cite{cbam} and ECA \cite{eca} both use a 2D convolution combined with channel attention.  The proposal of a dual attention network  \cite{dual} combining the channel, and spatial attention has gained popularity as a successful semantic segmentation method on 2D images. This method is highly successful but can not be replicated across many convolutions because of its high model complexity. 

Some attention mechanisms are applied to the 3D point cloud space with methods like SCA-CNN \cite{sca} and MPRM \cite{mrpm}. In these methods, spatial attention is utilized to extract contextual information across the spatial domain, while channel attention approaches are used for exploring local cross-channel interactions. We propose spatial and channel attention combinations to a 3D semantic segmentation network in attention-based weighting for kernel point convolution. Before kernel summation, an attention mechanism's introduction will allow the network to perform kernel weighting with considerations for factors outside of simple geometric relations. 

\section{Pyramid Point Network}
\subsection{Feature Pyramids}
Research in semantic segmentation in unorganized point clouds has mainly focused on the convolution unit of the network. Methods such as KPConv or ConvPoint \cite{convpoint} have proposed new approaches to performing convolutions on 3D points.  Even though the convolution units have changed, the network remains the same, following the basic U structure. We propose a new design of a multi-level network, which draws from the human perception module. The U network's idea is that as data progresses through the network, we can gain new information by varying the receptive fields. We look at the same scene multiple times from different levels. 

As the data progresses through the network, the encoder decreases the network's dimensionality while increasing the number of features; conversely, the decoder increases the number of dimensions while reducing the number of features. These different layers represent a mixture of high and low-level features. The initial higher dimensional layers display low-level features like edges, and the lower-dimensional features representing more complex and specific features. The issue with this approach is that decoder is sensitive to noise. Traversing the layers' dimensionality from high to low and back to high can contribute to a certain amount of noise, propagating with additional feature layers. 

We propose a change to combat these issues. Figure \ref{fig:arch} shows the full architecture of our multi-level focusing network.  The encoder-decoder structure stays the same, but the output is also upsampled in stages back to the original layer dimensionality for each decoding stage. We take a second look for each progressive layer, upsampling back to the actual network layers. The second look layers are then concatenated to the original decoder layer and processed. 

This approach has two distinct advantages. The first is that it provides multiple viewpoints at each layer, which are concatenated to make the final prediction. These additional viewpoints increase the network results. The other advantage is that these layers are not subject to the extra noise added as the features traverse through the network. These layers are less processed and, therefore, cleaner than their counterparts. 

Except for the first two downsampling layers, each layer is both downsampled and upsampled, resulting in several layers of similar features, each derived from a different point in the network. After some experimentation, we found the best configuration is to concatenate the similar elements at each layer of the network. In the next sections, we discuss the adjustments we made to the kernel point convolution and the encoder and decoder units in detail.

\subsection{Focused Kernel Point Convolution}
\begin{figure}[t]
\begin{center}
\includegraphics[width=0.99\linewidth]{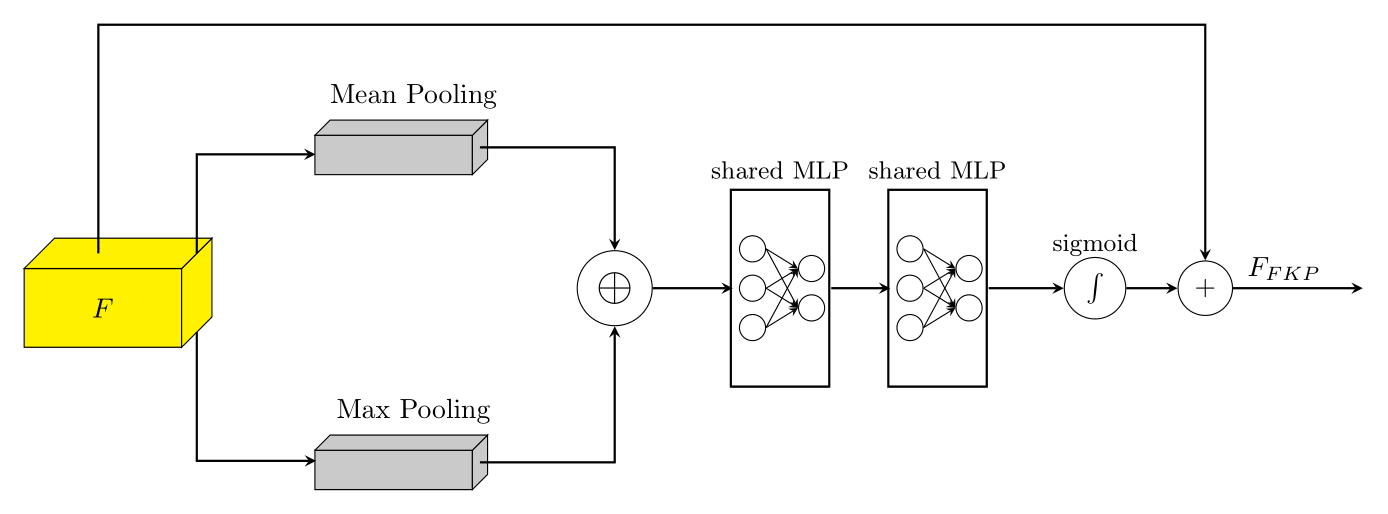}
 \caption{The kernel attention module. This module is applied to the kernel outputs in order to enhance the features, before reducing the kernel dimension.}
\label{fig:channelatten123}
\end{center}

\end{figure}
\subsubsection{Kernel Attention}
To address the idea of scale within different objects in a scene, we introduce the concept of kernel attention. In the original kernel point convolution solution, the final output features are produced by summing to the outputs of each of the $N$ number of kernels.  We argue that adding attention to the kernel modules can increase the network's discrimination ability. A direct linear correlation in the system may not always be the best application. By adding an attention mechanism, we can help discriminate which of the kernel weights has the most significant contribution. 

We propose an attention module on the kernel outputs before the final summation to produce the output features. This adjustment will allow the network to learn discrimination between the different kernel points and allow for a more sophisticated kernel weighting, rather than relying strictly on geometric correlation alone. The kernel attention is as follows:

We consider the kernel outputs as $F \in \mathbb{R}^{K\times N\times D}$, where $K$ is the number of kernel points, $N$ is the number of points and $D$ is the feature dimension. We take the max and mean pooling of the feature $F$, with respect to the kernel outputs, to get $F_{max} \in \mathbb{R}^{1 \times N \times D}$ and $F_{mean} \in \mathbb{R}^{1 \times N \times D}$ . $F_{mean}$ and $F_{max}$ are concatenated together and run through two shared MLPs. The MLP layers are followed by a sigmoid function, $\sigma$, which makes the kernel attention matrix $M_K(F) \in \mathbb{R}^{1\times N \times D}$. In the final step, we use element-wise multiplication to get the final feature $F'$. The entire process is shown below:

\begin{equation}
\begin{aligned}
M_K(F) = \sigma(MLP([F_{mean}, F_{max}])) \\
F_{FKP} = F \otimes M_K(F)
\end{aligned}
\end{equation}

A summation across all kernel features then reduces the final enhanced kernel outputs to make the final point features from our Focused Kernel Point Convolution, $F_{FKP} \in \mathbb{R}^{N \times D}$. The entire kernel attention architecture can be seen in Figure \ref{fig:channelatten123}. 

\begin{figure}[t]
\begin{center}
\includegraphics[width=0.99\linewidth]{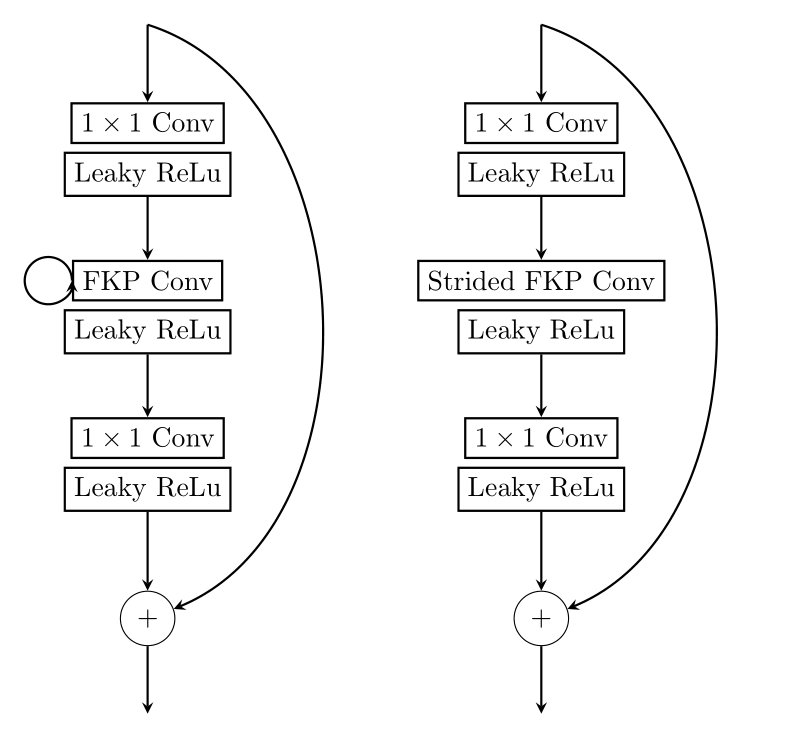}
\end{center}
   \caption{Example of our recurrent Focused Kernel Point Convolution bottleneck block and our Strided Focused Kernel Point Convolution bottleneck block. The traditional 2D convolution is replaced by a recurrent Focused Kernel Point Convolution and Focused Kernel Point Convolution, respectively}
\label{fig:bottleneck123}
\end{figure}

\begin{figure}[t]
\begin{center}
\includegraphics[width=0.95\linewidth]{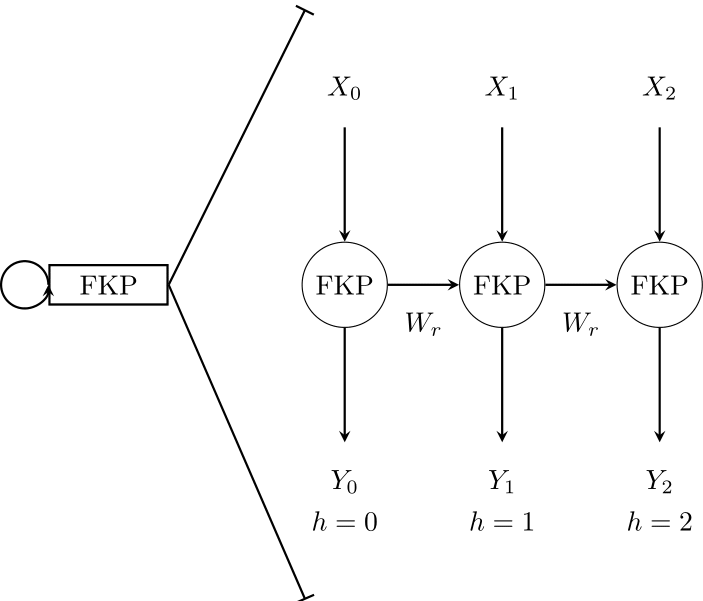}
\end{center}
   \caption{Example of the recurrent Focused Kernel Point Convolution, with three hidden layers.}
\label{fig:rfKP}
\end{figure}

\subsection{Encoder and Decoder}
Once we have established our FKP Conv, we then define our encoder's building blocks. We use the bottleneck structure commonly found in ResNet, with a few adjustments. We begin with a unary convolution, with a Leaky ReLu activation. This result is passed into a recurrent FKP Conv. We design the output of the recurrent layer as follows:

\begin{equation}
    \begin{aligned}
    O(t) = w_k^f * x^f(t) + w_k^r*x^r(t+1)
    \end{aligned}
\end{equation}

We can extend the equation to include any number of hidden layers. This recurrent FKP is followed by a Leaky ReLu activation and another unary convolution with Leaky ReLu activation. The final result is then summed with the original input features to form the final enhanced output. Figure \ref{fig:rfKP} shows an example of the deconstructed recurrent FKP Conv layout with three hidden layers. 

Our encoder unit consists of two layers; the first is the recurrent FKP bottleneck block followed by an FKP strided bottleneck block. This FKP strided bottleneck uses a single strided FKP Conv in place of the recurrent FKP Conv, as shown in Figure \ref{fig:bottleneck123}.  Our decoder consists of a nearest upsampling layer, followed by a unary convolution.

\section{Evaluations}
\begin{table*}[htbp]
\begin{center}
\begin{tabular}{lcccccccccc}
\hline
Method & mIoU (\%) & OA (\%) & \textit{ground}& \textit{buildings}&\textit{cars}&\textit{trucks}&\textit{poles}&\textit{power lines}&\textit{fences}&\textit{veg}      \\
\hline
ShellNet \cite{shellnet}&57.4&96.4&96.0&95.4&32.2&39.6&20.0&27.4&60.0&88.4\\
PointCNN \cite{pointcnn}&58.4&97.2&97.5&95.7&40.6&4.8&57.6&26.7&52.6&91.7\\
SuperPoint \cite{superpoint}&60.6&95.5&94.7&93.4&62.9&18.7&28.5&65.2&33.6&87.9\\
ConvPoint \cite{convpoint}&67.4&97.2&96.9&96.3&75.5&21.7&40.3&86.7&29.6&91.9\\
PointNet++ \cite{pointnet++}&68.3&95.7&94.1&89.1&75.4&30.3&40.0&79.9&46.2&91.2\\
Wang \cite{wang2020hierarchical}&72.6&94.6&95.8&95.2&86.3&\textbf{54.2}&43.6&92.4&58.9&94.1\\
KPConv \cite{kpconv}&81.1&97.8&97.1&96.6&85.3&41.9&75.0&95.5&63.5&94.1\\
\hline
Pyramid Point &	\textbf{83.6}&\textbf{98.3}&\textbf{97.8}&\textbf{97.3}&\textbf{88.4}&47.9&\textbf{77.6}&\textbf{96.7}&\textbf{67.5}&\textbf{95.4}\\
\hline
\end{tabular}
\caption{Current leaders on the DALES data set. We report the overall accuracy, mean IoU and per class IoU, for each category. We show that our network outperforms all other methods on the DALES data set. }
\label{tab:dales-results}
\end{center}
\end{table*}

\begin{figure*}[t]
\begin{center}
\includegraphics[width=0.33\linewidth]{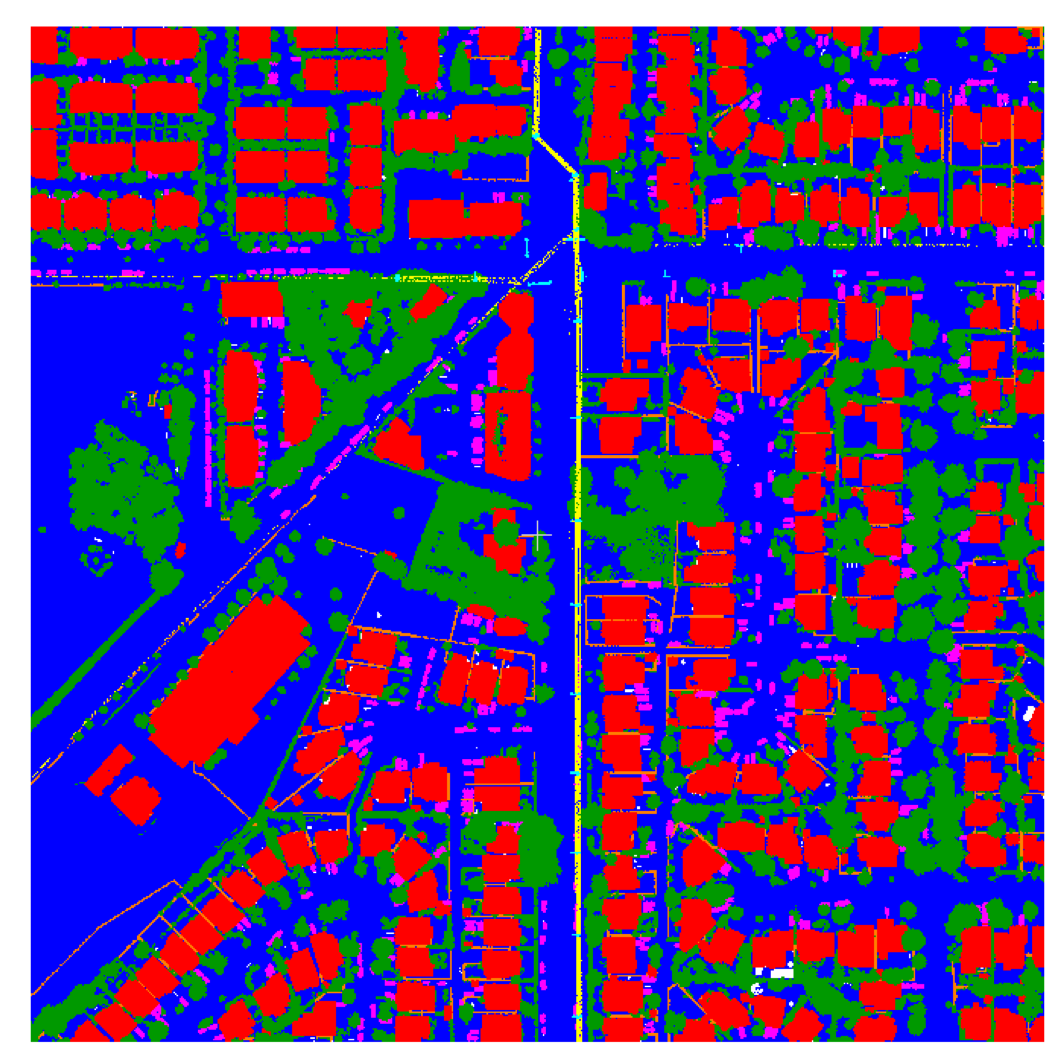}
\includegraphics[width=0.33\linewidth]{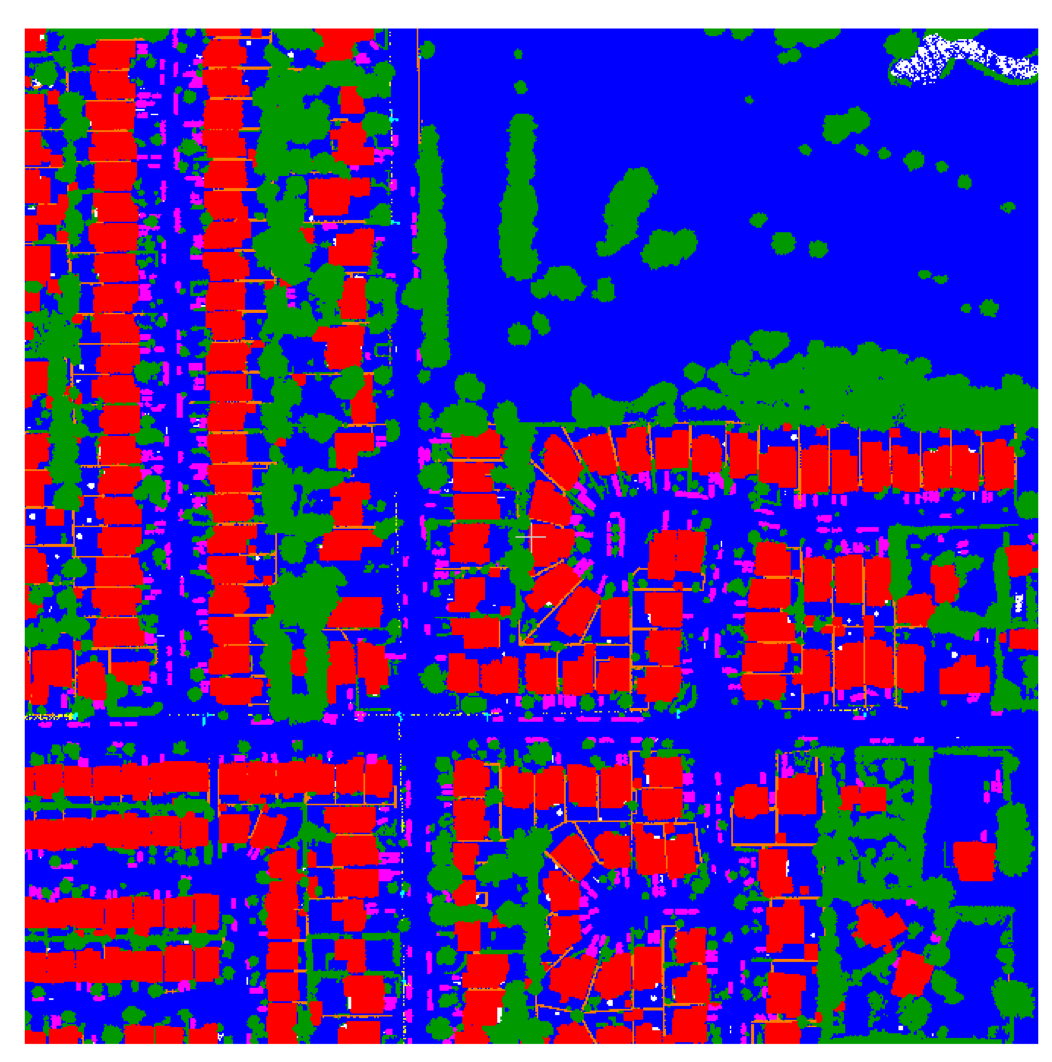}
\includegraphics[width=0.33\linewidth]{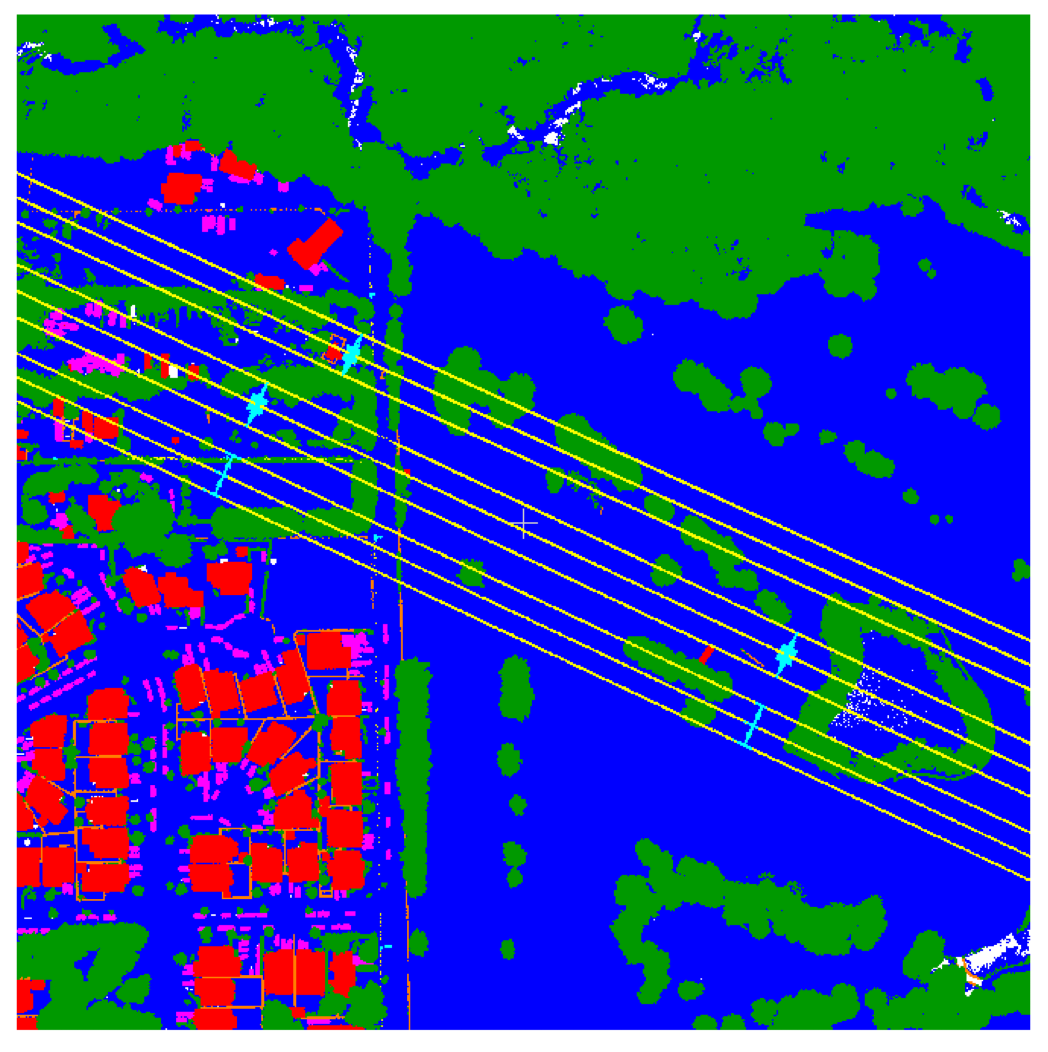}
   \caption{Visual results of the Pyramid Point network on the DALES data set. Points are colored by object category; ground (blue), buildings (red), cars (magenta), trucks (light orange), poles (cyan), power lines (yellow), fences (orange) and vegetation (green) }
   \label{fig:dales-vis}
   \end{center}
\end{figure*}

\begin{figure*}[t]
\begin{center}
\includegraphics[width=0.33\linewidth]{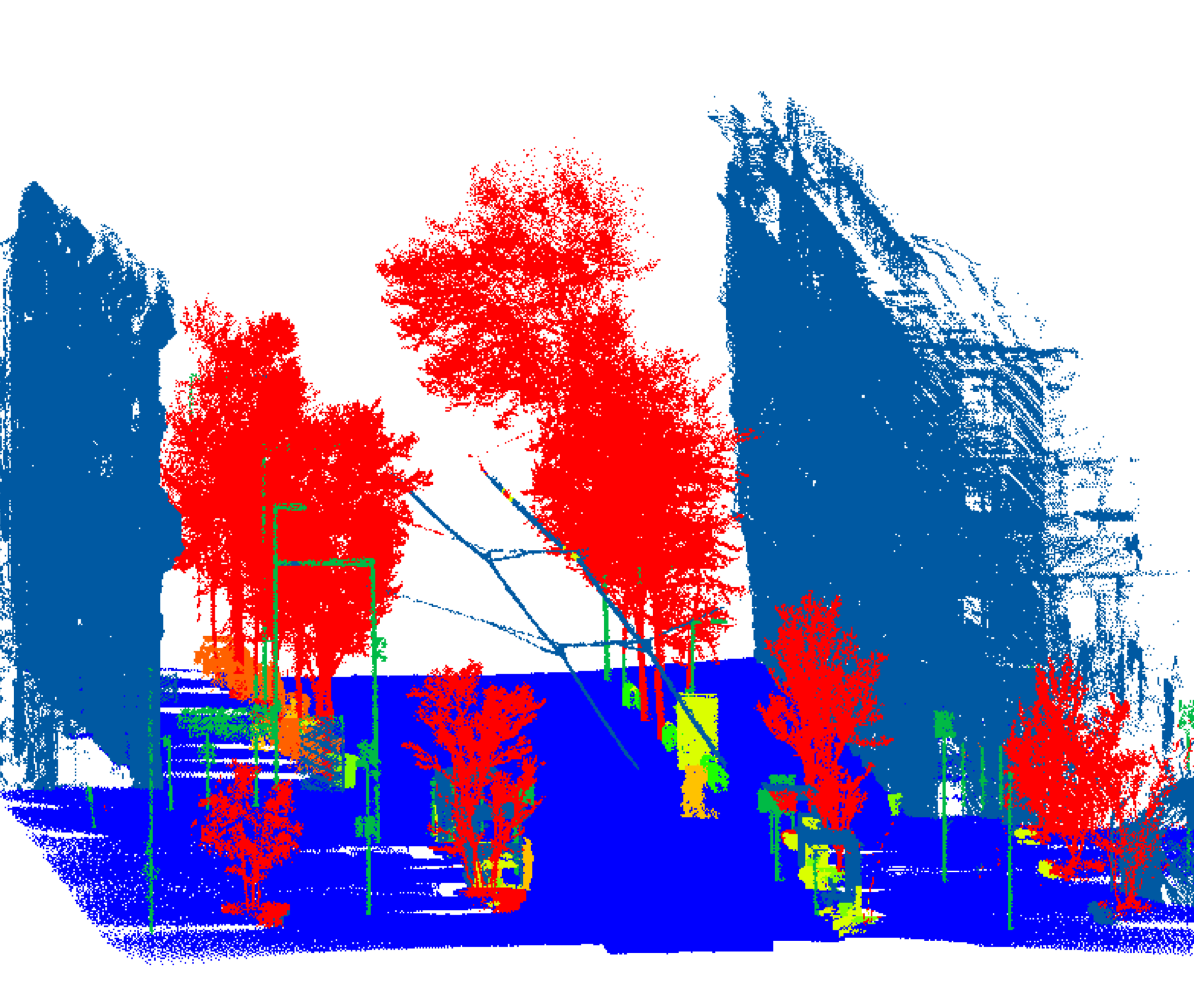}
\includegraphics[width=0.33\linewidth]{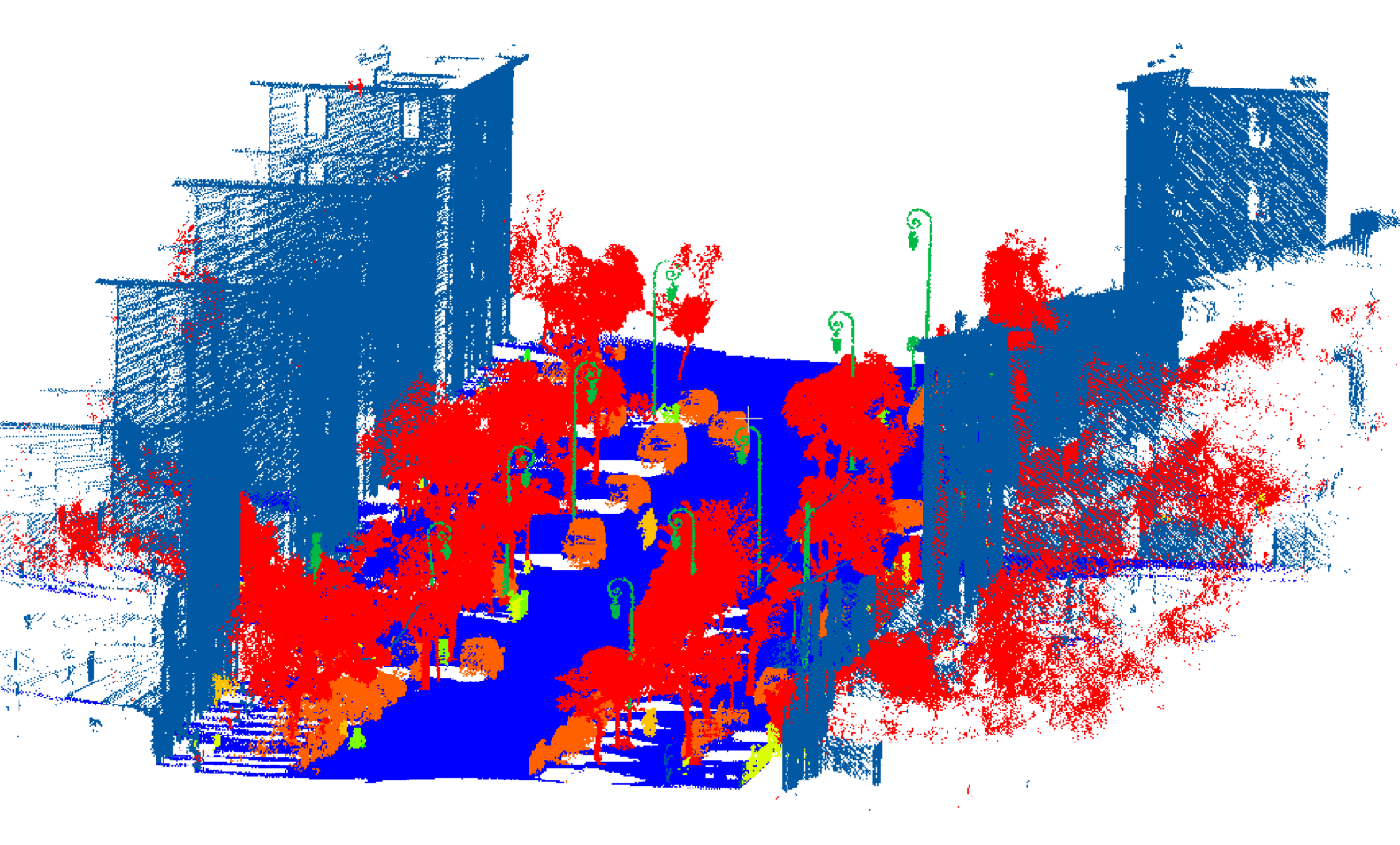}
\includegraphics[width=0.33\linewidth]{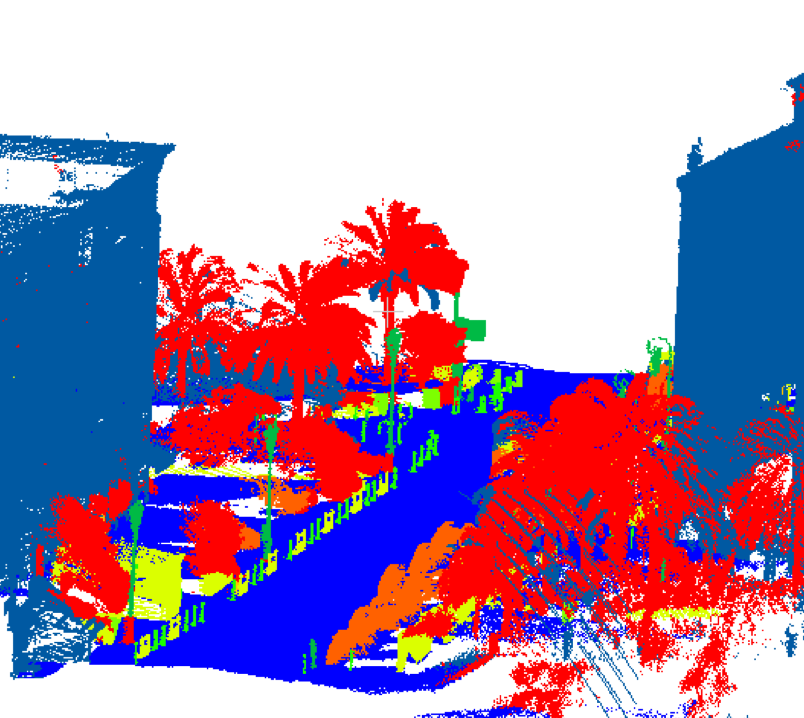}
   \caption{Visual results for the Paris-Lille 3D benchmark data set. Each point is colored by object category; ground (dark blue), buildings (light blue), pole (dark green), bollard (green), trash can (light green), barrier (yellow), pedestrian (light orange), car (dark orange) and natural (red)}
   \label{fig:npm3d-vis}
   \end{center}
\end{figure*}

\begin{table*}[htbp]
\begin{center}
\setlength{\tabcolsep}{5pt}
\begin{tabular}{lcccccccccc}
\hline
Method&mIoU(\%)&ground&building&pole&bollard&trash can&barrier&pedestrian&car&natural\\
\hline
ConvPoint \cite{convpoint}&75.9&99.5&95.1&71.6&88.7&46.7&52.9&53.5&89.4&85.4\\
RandLANet \cite{randla}&78.5&99.5&97.0&71.0&86.7&50.5&65.5&49.1&95.3&91.7\\
MS-RRFSegNet \cite{msrrfsegnet}&79.2&98.6&98.0&\textbf{79.7}&74.3&75.1&57.9&55.9&82.0&91.4\\
KP-FCNN \cite{kpconv}&82.0&99.5&94.0&71.3&83.1&\textbf{78.7}&47.7&78.2&94.4&91.4\\
FKAConv \cite{fkaconv}&82.7&\textbf{99.6}&\textbf{98.1}&77.2&\textbf{91.1}&64.7&\textbf{66.5}&58.1&\textbf{95.6}&\textbf{93.9}\\
\hline
Pyramid Point&\textbf{82.9}&\textbf{99.6}&97.1&74.6&84.3&56.0&65.9&\textbf{79.1}&95.1&\textbf{93.9}\\
\hline
\end{tabular}
\caption{Benchmark results on the Paris-Lille data set. Our method comes in first overall, with the best mean IoU. We also place first in the in the ground, natural and pedestrian categories. Accessed November 16\textsuperscript{th} 2020}
\label{tab:results-paris}
\end{center}
\end{table*}
We evaluate our network on three different types of LiDAR benchmark data sets; DALES \cite{dales}, Semantic3D \cite{semantic3d}, and Paris-Lille 3D \cite{npm3d}. We choose these benchmarks because of their popularity and because they represent three distinct types of LiDAR, Aerial Laser Scanners (ALS), Terrestrial Laser Scanners (TLS), and Mobile Laser Scanners (MLS). These different sensors all provide various object classes, resolutions, and types of occlusions and will allow us to test our network's robustness. 

 For each evaluation, we used five total layers in our architecture and began upsampling and downsampling simultaneously at layer three. The feature dimensions are $64/128/256/512/1024$, respectively. For our recurrent FKP Conv bottleneck, we use three hidden layers. After several experiments, we concluded that the best configuration for combining similarly shaped layers was to concatenate them. We changed the subsampling parameter and neighborhood radii based on the resolution of the point cloud. Aside from the subsampling rate and neighborhood radius, no other parameter changes were made when testing different data sets. 
 
 We examine our network's performance using mean Intersection over Union (IoU) as our primary metric. For consistency with the benchmark, we also report per class IoU and overall accuracy, where appropriate. A discussion of the results from each data set is below.
\subsection{DALES}
We evaluate the performance of our network on the DALES data set. DALES is a large-scale semantic segmentation data set for aerial LiDAR, containing over 500 million points and eight object categories in rural, suburban, industrial and commercial scenes. We used a subsampling rate of 0.25 meters and a neighborhood radius of 15 meters. 

Table \ref{tab:dales-results} shows the performance of our network on DALES. We can see that the Pyramid Point has a good performance with the highest mean IoU, of over 2.5\% more than the next closest method. One notable performance is between our network and KPConv, whose kernel point convolution makes up the backbone of our FKP Conv. We can see a large margin of performance improvement, especially in some of the lower performing categories. This emphasizes our pyramid architecture's effectiveness and ability to exploit the features' interconnections and bridge the performance gap between object categories. The visual results of some sample DALES scenes are shown in Figure \ref{fig:dales-vis}.

\subsection{Paris-Lille 3D}
We also test our method on the Paris Lille 3D data set. This data set is a mobile LiDAR data set covering two different French cities; the training and evaluation have ten labeled classes, covering around 2 km and 160 million points total, providing diverse, challenging scenes. We used a subsampling rate of 0.08 meters and a neighborhood radius of 4 meters. We compare our performance to other methods and measure our success by comparing the mean IoU across all different classes. 
Table \ref{tab:results-paris} shows our results compared against others in the benchmark. Overall, our method comes in first in this data set. We also claim the top performance in the ground, natural and pedestrian categories. We can also examine the qualitative results by looking at a selection of scenes shown in Figure \ref{fig:npm3d-vis}.

\begin{figure*}[t]
\begin{center}
\includegraphics[width=0.33\linewidth]{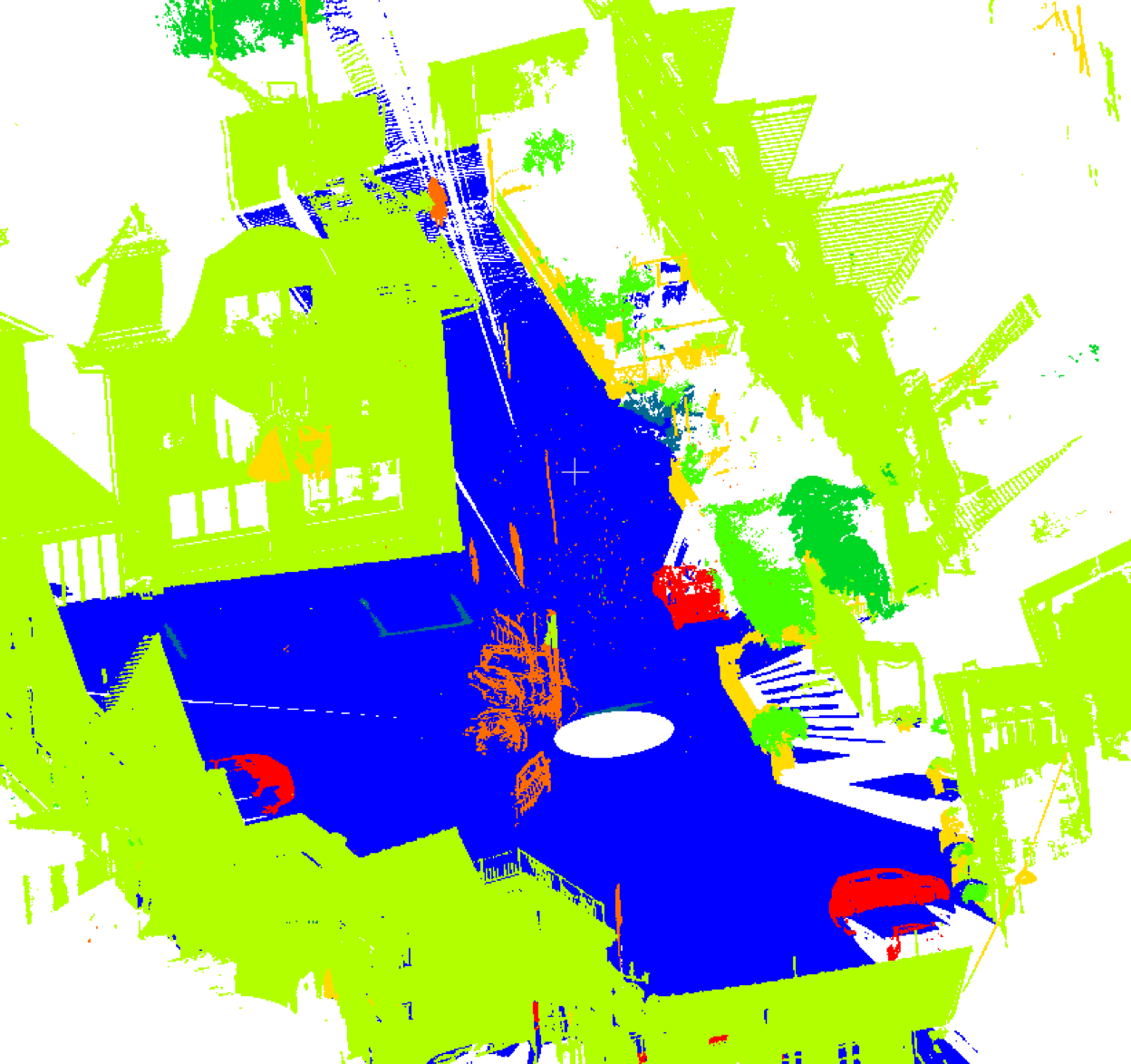}
\includegraphics[width=0.33\linewidth]{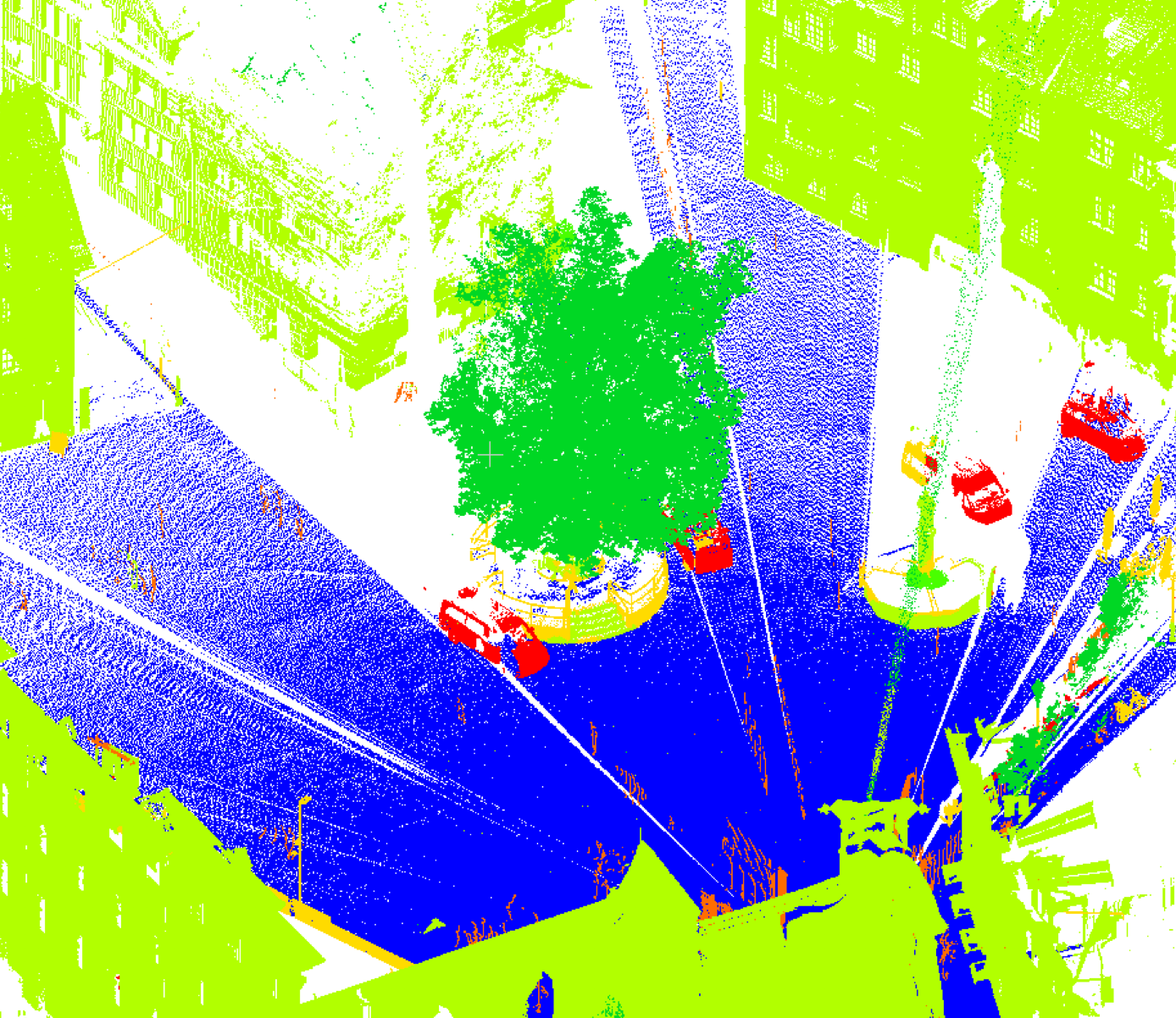}
\includegraphics[width=0.33\linewidth]{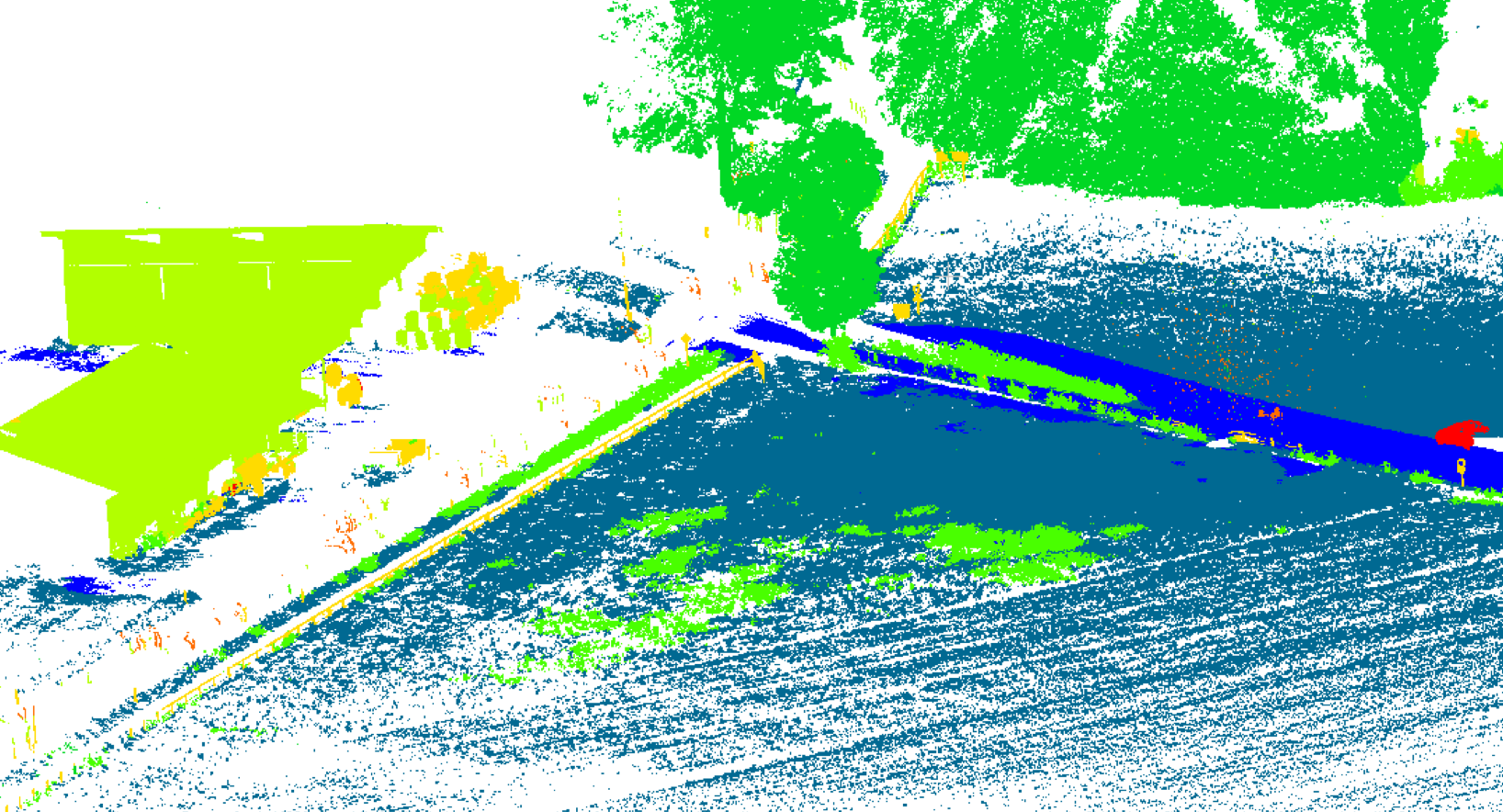}
   \caption{Visual results from the Semantic 3D benchmark data set. Points are colored by semantic class; man-made (orange), natural (dark green), , high veg (green), low veg (blue-green), buildings (light green), hardscape (yellow), scanning art (orange), cars (red)}
   \label{fig:sem-vis}
   \end{center}
\end{figure*}

\begin{table*}[htbp]
\centering
\setlength{\tabcolsep}{2pt}
\begin{tabular}{lcccccccccc}
\hline
Method & mIoU (\%) & OA (\%) & man-made & natural&high veg&low veg&buildings&hardscape & scanning art&cars\\
\hline
MSDeepVoxNet \cite{msdeep}&65.3&88.4&83.0&67.2&83.8&36.7&92.4&31.3&50.0&78.2\\
ShellNet \cite{shellnet}&69.3&93.2&96.3&90.4&83.9&41.0&94.2&34.7&43.9&70.2\\
GACNet \cite{GACNet}&70.8&91.9&86.4&77.7&\textbf{88.5}&\textbf{60.6}&94.2&37.3&43.5&77.8\\
SPG \cite{superpoint}&73.2&94.0&\textbf{97.4}&\textbf{92.6}&87.9&44.0&83.2&31.0&63.5&76.2\\
KPConv \cite{kpconv}&74.6&92.9&90.9&82.2&84.2&47.9&94.9&40.0&\textbf{77.3}&79.7\\
RandLA-Net \cite{randla}&\textbf{77.4}&\textbf{94.8}&95.6&91.4&86.6&51.5&\textbf{95.7}&\textbf{51.5}&69.8&76.8\\
\hline
Pyramid Point & 77.3&94.5&95.7&90.3&84.4&50.5&95.4&	45.9&72.7&\textbf{83.2}\\
\hline
\end{tabular}
\caption{ Benchmark results for the reduced-8 Semantic 3D data set. Our method comes in second overall of the available published results. Accessed November 16\textsuperscript{th} 2020}
\label{tab:sem3d-results}
\end{table*}
\begin{table}[h!]
\centering
\begin{tabular}{lcccccccccc}
\hline
\# Hidden Layers&mIoU\\
\hline
2 & 75.6	\\
3 & 77.3	\\
4 & 76.6 	\\
\hline
\end{tabular}
\caption{Ablation study of the effect of residual layers on the Semantic3D data set }
\label{tab:ablation}
\end{table}

\begin{table}[h!]
\centering
\begin{tabular}{lcccccccccc}
\hline
Variation &mIoU\\
\hline
(1) No Focused Kernel & 85.4 \\
(2) Max Focused Kernel & 89.1 \\
(3) Mean Focused Kernel & 88.3 \\
(4) Max, Mean Focused Kernel & 90.2 	\\
\hline
\end{tabular}
\caption{Ablation study on the combination of intermediate layers in the Pyramid Point network}
\label{tab:ablation-layers}
\end{table}

\subsection{Semantic3D}
\setlength{\parskip}{-1pt}
The third point cloud classification evaluation is on the Semantic3D data set. This data set is taken with a terrestrial LiDAR sensor and has over four billion hand-labeled points with eight semantic classes. We used a subsampling rate of 0.06 meters and a neighborhood radius of 3 meters. We use the reduced-8 data set for our evaluations because we believe that the sampling strategy is less biased towards the points closest to the sensor. \\
This data set is challenging because of its overall size, occlusion types, and varying resolutions. The results of our network are in Table \ref{tab:sem3d-results}. Like others \cite{randla, kpconv}; we only compare published methods, and we obtain the scores from their benchmark. We can see that our method outperforms all other methods except for RandLA Net, which performs similarly with a 0.1\% difference between the two methods. Pyramid Point also logs the highest performance in the cars category, by 3.5\%. Figure \ref{fig:sem-vis} shows the qualitative results from selected scenes in Semantic3D. 
\subsection{Ablation Study}
We perform two ablation studies to discover the effects of some of our contributions to the network. We focus on the impact of the recurrent layers and how many hidden layers contribute to positive network performances. We also explore how the FKP Conv and variations to the pooling methods within FKP Conv can affect the final results. 
\subsubsection{Recurrent Layers}
 The first set of experiments compares the full intact network and observes the effects of different numbers of hidden layers in our recurrent structure.  We use Semantic3D data to perform this experiment. Table \ref{tab:ablation} shows the effects of different numbers of hidden layers. Adding the recurrent layers improves the performance, increasing when going from two to three hidden layers. This performance drops off when increasing the number of hidden layers from three to four. 

\subsubsection{FKP Conv}
Next, we compare the FKP Conv and how it affects our network results. We first examine the network's effects using a kernel point convolution, without the kernel attention elements. We explore using max and mean pooling in the attention mechanism, and then finally, we explore the combination of all of them to make up the final FKP Conv. For this evaluation, we use a subset of the Paris-Lille 3D data. 

Table \ref{tab:ablation-layers} compares the results from all ablated networks. We note that adding the kernel's attention to the kernel point convolution gives the biggest jump in performance; we then get a marginal improvement by considering the max and mean pooling together to form the final FKP Conv element. 

\section{Conclusion}
In this paper, we have presented a novel network for semantic segmentation in 3D point clouds. Our Pyramid Point network uses the concept of a dense pyramid structure to increase the number of receptive fields and revisit feature layers. This structure, which varies from the traditional U-style networks, provides additional details and less noise in the feature channels. \\
We also introduced a FKP Conv, which utilizes an attention mechanism to improve outputs from the convolutional kernel.  We demonstrated the network's success by showing results on three prominent benchmark data sets, DALES, Semantic3D, and Paris-Lille 3D. The Pyramid Point was the top performer in the DALES and Paris-Lille 3D data sets and performed competitively in the Semantic3D data set, demonstrating the network's success in three different types of LiDAR environments. We also performed an ablation study, which illustrates how each element we have introduced positively contributes to our network performance. 

\section*{Acknowledgments}
This effort was supported in part by the U.S. Air Force through contract number FA8650-20-F-1925.  The views expressed in this article are those of the authors and do not reflect on the official policy of the Air Force, Department of Defense or the U.S. Government.

\newpage
{\small
\bibliographystyle{ieee_fullname}
\bibliography{egbib}

\begin{thebibliography}{10}\itemsep=-1pt

\bibitem{nablan}
Md~Zahangir Alom, Theus Aspiras, Tarek~M Taha, and Vijayan~K Asari.
\newblock Skin cancer segmentation and classification with nabla-n and
  inception recurrent residual convolutional networks.
\newblock {\em arXiv preprint arXiv:1904.11126}, 2019.

\bibitem{r2unet}
Md~Zahangir Alom, Mahmudul Hasan, Chris Yakopcic, Tarek~M Taha, and Vijayan~K
  Asari.
\newblock Recurrent residual convolutional neural network based on u-net
  (r2u-net) for medical image segmentation.
\newblock {\em arXiv preprint arXiv:1802.06955}, 2018.

\bibitem{pcnn}
Matan Atzmon, Haggai Maron, and Yaron Lipman.
\newblock Point convolutional neural networks by extension operators.
\newblock {\em arXiv preprint arXiv:1803.10091}, 2018.

\bibitem{segnet}
Vijay Badrinarayanan, Alex Kendall, and Roberto Cipolla.
\newblock Segnet: A deep convolutional encoder-decoder architecture for image
  segmentation.
\newblock {\em IEEE transactions on pattern analysis and machine intelligence},
  39(12):2481--2495, 2017.

\bibitem{bay2006surf}
Herbert Bay, Tinne Tuytelaars, and Luc Van~Gool.
\newblock Surf: Speeded up robust features.
\newblock In {\em European conference on computer vision}, pages 404--417.
  Springer, 2006.

\bibitem{convpoint}
Alexandre Boulch.
\newblock Convpoint: Continuous convolutions for point cloud processing.
\newblock {\em Computers \& Graphics}, 2020.

\bibitem{fkaconv}
Alexandre Boulch, Gilles Puy, and Renaud Marlet.
\newblock Fkaconv: Feature-kernel alignment for point cloud convolution, 2020.

\bibitem{attentivesurvey}
Sneha Chaudhari, Gungor Polatkan, Rohan Ramanath, and Varun Mithal.
\newblock An attentive survey of attention models.
\newblock {\em arXiv preprint arXiv:1904.02874}, 2019.

\bibitem{sca}
Long Chen, Hanwang Zhang, Jun Xiao, Liqiang Nie, Jian Shao, Wei Liu, and
  Tat-Seng Chua.
\newblock Sca-cnn: Spatial and channel-wise attention in convolutional networks
  for image captioning.
\newblock In {\em Proceedings of the IEEE conference on computer vision and
  pattern recognition}, pages 5659--5667, 2017.

\bibitem{dual}
Jun Fu, Jing Liu, Haijie Tian, Yong Li, Yongjun Bao, Zhiwei Fang, and Hanqing
  Lu.
\newblock Dual attention network for scene segmentation.
\newblock In {\em Proceedings of the IEEE Conference on Computer Vision and
  Pattern Recognition}, pages 3146--3154, 2019.

\bibitem{flexconvolution}
Fabian Groh, Patrick Wieschollek, and Hendrik~PA Lensch.
\newblock Flex-convolution.
\newblock In {\em Asian Conference on Computer Vision}, pages 105--122.
  Springer, 2018.

\bibitem{semantic3d}
Timo Hackel, Nikolay Savinov, Lubor Ladicky, Jan~D Wegner, Konrad Schindler,
  and Marc Pollefeys.
\newblock Semantic3d. net: A new large-scale point cloud classification
  benchmark.
\newblock {\em arXiv preprint arXiv:1704.03847}, 2017.

\bibitem{randla}
Qingyong Hu, Bo Yang, Linhai Xie, Stefano Rosa, Yulan Guo, Zhihua Wang, Niki
  Trigoni, and Andrew Markham.
\newblock Randla-net: Efficient semantic segmentation of large-scale point
  clouds.
\newblock In {\em Proceedings of the IEEE/CVF Conference on Computer Vision and
  Pattern Recognition}, pages 11108--11117, 2020.

\bibitem{pointwise}
Binh-Son Hua, Minh-Khoi Tran, and Sai-Kit Yeung.
\newblock Pointwise convolutional neural networks.
\newblock In {\em Proceedings of the IEEE Conference on Computer Vision and
  Pattern Recognition}, pages 984--993, 2018.

\bibitem{interlaced}
Lang Huang, Yuhui Yuan, Jianyuan Guo, Chao Zhang, Xilin Chen, and Jingdong
  Wang.
\newblock Interlaced sparse self-attention for semantic segmentation.
\newblock {\em arXiv preprint arXiv:1907.12273}, 2019.

\bibitem{attnsurvey}
Asifullah Khan, Anabia Sohail, Umme Zahoora, and Aqsa~Saeed Qureshi.
\newblock A survey of the recent architectures of deep convolutional neural
  networks.
\newblock {\em Artificial Intelligence Review}, pages 1--62, 2020.

\bibitem{alexnet}
Alex Krizhevsky, Ilya Sutskever, and Geoffrey~E Hinton.
\newblock Imagenet classification with deep convolutional neural networks.
\newblock In {\em Advances in neural information processing systems}, pages
  1097--1105, 2012.

\bibitem{superpoint}
Loic Landrieu and Martin Simonovsky.
\newblock Large-scale point cloud semantic segmentation with superpoint graphs.
\newblock In {\em Proceedings of the IEEE Conference on Computer Vision and
  Pattern Recognition}, pages 4558--4567, 2018.

\bibitem{pointcnn}
Yangyan Li, Rui Bu, Mingchao Sun, Wei Wu, Xinhan Di, and Baoquan Chen.
\newblock Pointcnn: Convolution on x-transformed points.
\newblock In {\em Advances in neural information processing systems}, pages
  820--830, 2018.

\bibitem{pointvoxelcnn}
Zhijian Liu, Haotian Tang, Yujun Lin, and Song Han.
\newblock Point-voxel cnn for efficient 3d deep learning.
\newblock In {\em Advances in Neural Information Processing Systems}, pages
  965--975, 2019.

\bibitem{lowe2004sift}
G Lowe.
\newblock Sift-the scale invariant feature transform.
\newblock {\em Int. J}, 2:91--110, 2004.

\bibitem{msrrfsegnet}
Haifeng Luo, Chongcheng Chen, Lina Fang, Kourosh Khoshelham, and Guixi Shen.
\newblock Ms-rrfsegnet: Multiscale regional relation feature segmentation
  network for semantic segmentation of urban scene point clouds.
\newblock {\em IEEE Transactions on Geoscience and Remote Sensing}, 2020.

\bibitem{voxnet}
Daniel Maturana and Sebastian Scherer.
\newblock Voxnet: A 3d convolutional neural network for real-time object
  recognition.
\newblock In {\em 2015 IEEE/RSJ International Conference on Intelligent Robots
  and Systems (IROS)}, pages 922--928. IEEE, 2015.

\bibitem{votenet}
Charles~R Qi, Or Litany, Kaiming He, and Leonidas~J Guibas.
\newblock Deep hough voting for 3d object detection in point clouds.
\newblock In {\em Proceedings of the IEEE International Conference on Computer
  Vision}, pages 9277--9286, 2019.

\bibitem{pointnet}
Charles~R Qi, Hao Su, Kaichun Mo, and Leonidas~J Guibas.
\newblock Pointnet: Deep learning on point sets for 3d classification and
  segmentation.
\newblock In {\em Proceedings of the IEEE conference on computer vision and
  pattern recognition}, pages 652--660, 2017.

\bibitem{pointnet++}
Charles~Ruizhongtai Qi, Li Yi, Hao Su, and Leonidas~J Guibas.
\newblock Pointnet++: Deep hierarchical feature learning on point sets in a
  metric space.
\newblock In {\em Advances in neural information processing systems}, pages
  5099--5108, 2017.

\bibitem{unet}
Olaf Ronneberger, Philipp Fischer, and Thomas Brox.
\newblock U-net: Convolutional networks for biomedical image segmentation.
\newblock In {\em International Conference on Medical image computing and
  computer-assisted intervention}, pages 234--241. Springer, 2015.

\bibitem{msdeep}
Xavier Roynard, Jean-Emmanuel Deschaud, and Fran{\c{c}}ois Goulette.
\newblock Classification of point cloud scenes with multiscale voxel deep
  network.
\newblock {\em arXiv preprint arXiv:1804.03583}, 2018.

\bibitem{npm3d}
Xavier Roynard, Jean-Emmanuel Deschaud, and François Goulette.
\newblock Paris-lille-3d: A large and high-quality ground-truth urban point
  cloud dataset for automatic segmentation and classification.
\newblock {\em The International Journal of Robotics Research}, 37(6):545--557,
  2018.

\bibitem{splatnet}
Hang Su, Varun Jampani, Deqing Sun, Subhransu Maji, Evangelos Kalogerakis,
  Ming-Hsuan Yang, and Jan Kautz.
\newblock Splatnet: Sparse lattice networks for point cloud processing.
\newblock In {\em Proceedings of the IEEE Conference on Computer Vision and
  Pattern Recognition}, pages 2530--2539, 2018.

\bibitem{kpconv}
Hugues Thomas, Charles~R Qi, Jean-Emmanuel Deschaud, Beatriz Marcotegui,
  Fran{\c{c}}ois Goulette, and Leonidas~J Guibas.
\newblock Kpconv: Flexible and deformable convolution for point clouds.
\newblock In {\em Proceedings of the IEEE International Conference on Computer
  Vision}, pages 6411--6420, 2019.

\bibitem{dales}
Nina Varney, Vijayan~K Asari, and Quinn Graehling.
\newblock Dales: A large-scale aerial lidar data set for semantic segmentation.
\newblock In {\em Proceedings of the IEEE/CVF Conference on Computer Vision and
  Pattern Recognition Workshops}, pages 186--187, 2020.

\bibitem{GACNet}
Lei Wang, Yuchun Huang, Yaolin Hou, Shenman Zhang, and Jie Shan.
\newblock Graph attention convolution for point cloud semantic segmentation.
\newblock In {\em Proceedings of the IEEE Conference on Computer Vision and
  Pattern Recognition}, pages 10296--10305, 2019.

\bibitem{eca}
Qilong Wang, Banggu Wu, Pengfei Zhu, Peihua Li, Wangmeng Zuo, and Qinghua Hu.
\newblock Eca-net: Efficient channel attention for deep convolutional neural
  networks.
\newblock In {\em Proceedings of the IEEE/CVF Conference on Computer Vision and
  Pattern Recognition}, pages 11534--11542, 2020.

\bibitem{wang2020hierarchical}
Yongjun Wang, Tengping Jiang, Jing Liu, Xiaorui Li, and Chong Liang.
\newblock Hierarchical instance recognition of individual roadside trees in
  environmentally complex urban areas from uav laser scanning point clouds.
\newblock {\em ISPRS International Journal of Geo-Information}, 9(10):595,
  2020.

\bibitem{mrpm}
Jiacheng Wei, Guosheng Lin, Kim-Hui Yap, Tzu-Yi Hung, and Lihua Xie.
\newblock Multi-path region mining for weakly supervised 3d semantic
  segmentation on point clouds.
\newblock In {\em Proceedings of the IEEE/CVF Conference on Computer Vision and
  Pattern Recognition}, pages 4384--4393, 2020.

\bibitem{cbam}
Sanghyun Woo, Jongchan Park, Joon-Young Lee, and In So~Kweon.
\newblock Cbam: Convolutional block attention module.
\newblock In {\em Proceedings of the European conference on computer vision
  (ECCV)}, pages 3--19, 2018.

\bibitem{spidercnn}
Yifan Xu, Tianqi Fan, Mingye Xu, Long Zeng, and Yu Qiao.
\newblock Spidercnn: Deep learning on point sets with parameterized
  convolutional filters.
\newblock In {\em Proceedings of the European Conference on Computer Vision
  (ECCV)}, pages 87--102, 2018.

\bibitem{rcan}
Yulun Zhang, Kunpeng Li, Kai Li, Lichen Wang, Bineng Zhong, and Yun Fu.
\newblock Image super-resolution using very deep residual channel attention
  networks.
\newblock In {\em Proceedings of the European Conference on Computer Vision
  (ECCV)}, pages 286--301, 2018.

\bibitem{shellnet}
Zhiyuan Zhang, Binh-Son Hua, and Sai-Kit Yeung.
\newblock Shellnet: Efficient point cloud convolutional neural networks using
  concentric shells statistics.
\newblock In {\em Proceedings of the IEEE International Conference on Computer
  Vision}, pages 1607--1616, 2019.

\bibitem{voxelnet}
Yin Zhou and Oncel Tuzel.
\newblock Voxelnet: End-to-end learning for point cloud based 3d object
  detection.
\newblock In {\em Proceedings of the IEEE Conference on Computer Vision and
  Pattern Recognition}, pages 4490--4499, 2018.

\end{thebibliography}
}

\end{document}